\documentclass[conference,letterpaper, 10pt]{ieeeconf}
\pdfoutput=1
\IEEEoverridecommandlockouts

\usepackage{macros}
\usepackage[backend=biber,style=ieee,natbib=true,doi=false,isbn=false,url=false,mincitenames=1]{biblatex} 
\usepackage{amsmath,amssymb,amsfonts}
\usepackage{algorithmic}
\usepackage{graphicx}
\usepackage{textcomp}
\usepackage{color}
\usepackage{caption}
\usepackage{subcaption}
\usepackage{cleveref}
\crefname{algocf}{alg.}{algs.}
\Crefname{algocf}{Algorithm}{Algorithms}
\usepackage{amsmath}
\usepackage{amssymb}
\usepackage{fixmath}
\usepackage{siunitx}
\usepackage[noend]{algorithm2e}
\SetAlFnt{\footnotesize}
 \RestyleAlgo{ruled}
\def\BibTeX{{\rm B\kern-.05em{\sc i\kern-.025em b}\kern-.08em
    T\kern-.1667em\lower.7ex\hbox{E}\kern-.125emX}}
\PassOptionsToPackage{hyphens}{url}
\usepackage{xcolor}

\SetCommentSty{mycommfont}

\showirosfalse
\showarxivtrue

\begin{document}
\title{Fast and Scalable Signal Inference for Active Robotic Source Seeking\\
\thanks{\textsuperscript{1} NASA Jet Propulsion Laboratory, Caltech, \textsuperscript{2} University of Southern California, \textsuperscript{3} Stanford University. Corresponding Author: cdennist@usc.edu. Sukhatme holds concurrent appointments as a Professor at USC and as an Amazon Scholar. This paper describes work not associated with Amazon.}
\thanks{The work is partially supported by the Jet Propulsion Laboratory, California Institute of Technology, under a contract with the National Aeronautics and Space Administration (80NM0018D0004), and Defense Advanced Research Projects Agency (DARPA). ©2022. All rights reserved. 
}
}

\author{Christopher E. Denniston\textsuperscript{1,2}
\and
Oriana Peltzer\textsuperscript{3}
\and
Joshua Ott\textsuperscript{3}
\and 
Sangwoo Moon\textsuperscript{1}
\and
Sung-Kyun Kim\textsuperscript{1}
\and 
Gaurav S. Sukhatme\textsuperscript{2}
\and 
Mykel J. Kochenderfer\textsuperscript{3}
\and 
Mac Schwager\textsuperscript{3}
\and 
Ali-akbar Agha-mohammadi\textsuperscript{1}
}

\maketitle

\begin{abstract}
In active source seeking, a robot takes repeated measurements 
in order to locate a signal source in a cluttered and unknown environment.
A key component of an active source seeking robot planner is a model that can produce estimates of the signal at unknown locations with uncertainty quantification.
This model allows the robot to plan for future measurements in the environment.
Traditionally, this model has been in the form of a Gaussian process, which has difficulty scaling and cannot represent obstacles.
We propose a global and local factor graph model for active source seeking, which allows the model to scale to a large number of measurements and represent unknown obstacles in the environment.
We combine this model with extensions to a highly scalable planner to form a system for large-scale active source seeking.
We demonstrate that our approach outperforms baseline methods in both simulated and real robot experiments.
\end{abstract}

\section{Introduction}

Prompt and accurate situational awareness is crucial for first responders to emergencies, such as natural disasters or accidents. 
Visual information is often unavailable or insufficient to detect people needing assistance in disaster-stricken environments.
Wireless signals, such as radio, Wi-Fi, or Bluetooth from mobile devices, can be important in search and rescue missions.
Autonomous exploration and signal source seeking by a robotic system can significantly improve the situational awareness of the human response team operating in hazardous and extreme environments.
In active source seeking, a robot not only attempts to get closer to the predicted source location but also gathers more information to actively reduce the source localization uncertainty.

Signal source seeking in large, unknown, obstacle-rich environments is an important but challenging task.
Source seeking systems typically consist of two components: a model of the environment that provides a belief over the signal strength at unknown locations, and a planner which generates a policy to improve the model and locate the source.
The model of the environment, as well as the planner, must scale to a large area and a large number of collected measurements.
Models for signal inference typically do not scale well to large numbers of measurements, relying on either approximation or having exponential growth in the time required for inference~\cite{kemna_voronoi,mccammon_ocean_2021,Rasmussen2006}.
In the presence of unknown obstacles, the robot must continually re-plan at a high rate and adapt the signal model to account for newly discovered obstacles.
The presence of unknown obstacles has complex impacts on signal propagation and requires the robot have the ability to re-visit un-visited corridors which potentially lead to the signal source.
\begin{figure}
    \centering
    \includegraphics[width=\columnwidth]{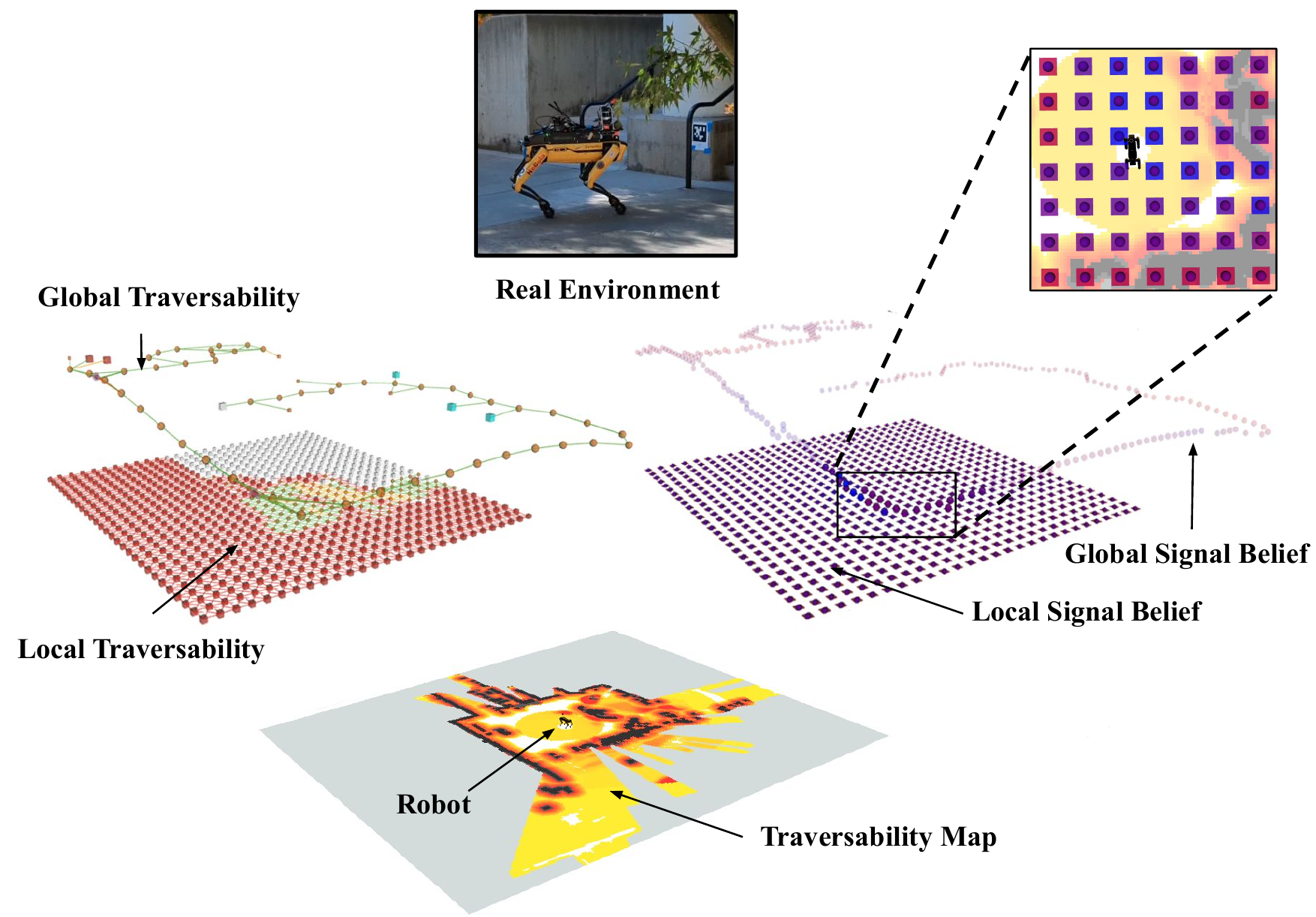}
    \caption{{\bf Overview of approach.} Both the information roadmap (which models the traversabilty of the environment) and the signal inference system maintain global and local representations of the environment.
    The signal inference system topology is built from the information roadmap. 
      The environment is a 
    paved outdoor walkway
    which can be seen in the image as well as the traversability map. 
    The local signal representation maintains both the mean (shown as colored spheres)  as well as the variance (flat squares), both are shown from blue (low) to red (high).}
    \label{fig:hero}
    \vspace{-1.5em}
\end{figure}

In this work, we formulate the problem of finding the highest signal strength as a problem of finding the location with the maximum radios signal to noise ratio (SNR). 
To find this maximum, the robot takes actions which increase its understanding of the signal while the planner is creating the next policy for the robot to execute. 
We formulate this as an Bayesian informative path planning problem in which the robot must balance the exploration-exploitation trade off~\cite{marchant_sequential_2014}. 
In order to find the signal source, the robot must explore to better the model of the signal in the environment, while being driven to the maxima of the signal strength.

To improve understanding of the signal strength, the robot must be able to estimate both the strength at unknown spatial locations, as well as the uncertainty in that estimate.
This probabilistic signal model gives the robot the ability to plan over future measurements to determine which locations to measure given the information about the signal strength.

To facilitate large-scale active source seeking both the planner and model of the environment must be highly scalable. 
To this end, we have developed a factor-graph based signal inference model and extended a large-scale multi-robot planner to handle these challenges.
Both the planner and model maintain global and local representations of the environment, which allows for decomposition of the problem.
This problem decomposition can be seen in \cref{fig:hero}, which shows that both the information roadmap (which contains traversibility information) and the signal inference are split into global and local components.

Our contributions are:
\begin{itemize}
    \item A novel factor graph based model for belief over the signal strength which scales well with the number of measurements taken and models the signal propagation through obstacles,
    \item An extension to the PLGRIM planner~\cite{kim2021plgrim} that allows it to take advantage of our factor-graph model and perform active source seeking, and
    \item Demonstration on challenging simulation and real-robot experiments with real-world signal strength data.
\end{itemize}

\section{Background / Related Work}
\textbf{Active Source seeking} is the process of using a robot to find a signal source in an unknown environment. 
Active source seeking has been used in locating radiation sources by representing the environment as voxels~\cite{icra_voxel2018}.
Active source seeking has also been used to map plumes of unknown hazardous agents~\cite{yoonchang_online_nodate}.
Gaussian processes have been used as the model in active source seeking of radio signals by developing a control law that allows the robot to move towards the source, while updating the model in an online fashion~\cite{online_radio}.
Active source seeking has been extended to the multi-robot setting by using a particle filter based model~\cite{Charrow2012CooperativeME}.

\textbf{Informative path planning} (IPP) is a process in which a robot takes measurements of a concentration to maximize an information metric.
Informative path planning can be used to minimize the uncertainty in the model, such as through an entropy metric~\cite{g2005}.
Informative path planning can also be combined with Bayesian acquisition funtions to form sequential Bayesian optimization~\cite{marchant_sequential_2014}, in which the goal is to find high concentration areas and can be used for active source seeking.
Previously, this formulation has been used in finding high concentrations of chlorophyll for studying algae blooms~\cite{mccammon_ocean_2021,fossum_information-driven_nodate} or finding the quantiles of a chlorophyll distribution~\cite{denniston_ral}.

\textbf{Gaussian Processes} are non-parametric models with uncertainty quantification which are widely used for IPP~\cite{marchant_sequential_2014,denniston_ral,kemna_voronoi,online_radio,denniston_icra_2021}.
They approximate an unknown function from its known outputs by computing the similarity between points from a kernel function, $k$, in our case the squared exponential kernel~\cite{Rasmussen2006}.
Function values $y_*$ at any input location $\mathbf{x}_*$ are approximated by a Gaussian distribution:
 $\sensedvalues_* | \sensedvalues \sim \mathcal{N} \left(K_* K^{-1} y,~ K_{**} - K_* K^{-1} K^T_* \right) $
where
 $\sensedvalues$ is training output,
 $\sensedvalues_*$ is test output.
The kernel matrices $K$, $K_*$, and $K_{**}$ are computed by evaluating the kernel on $\sensedlocations$, the training input, and
 $\sensedlocations_*$, the test input, such that $K^{i,j} = k(x^i,x^j)$, $K_*^{i,j} = k(x^i, x_{*}^j)$, and $K_{**}^{i,j} = k(\sensedlocation_{*}^i, \sensedlocation_{*}^j)$.

Gaussian processes typically do not have a way to represent obstacles as the kernel function only relies only on the distance between the locations in the model.
Gaussian processes tend to scale cubically  in the time required for inference due to the need to compute $K^{-1}$, and scale linearly in the time required when adding new measurements as the kernel function $k$ needs to be computed with the new measurement and all previous measurements~\cite{Rasmussen2006}.

\textbf{Ergodic trajectory generation }is rooted in the intuition that a path should spend time in a region proportional to the amount of expected information in that region \cite{dressel2017efficient, dressel2018optimality}. Ergodic trajectory design was originally presented in \citet{mathew2011metrics} where they introduced a norm on the statistical distance between a trajectory and a reference distribution allowing problem to be framed as an optimization problem with the goal of achieving the lowest ergodic score. 
\citet{miller2015optimal} introduced a closed-loop ergodic control algorithm for active search problems and \citet{dressel2018efficient} extended this work to target source localization. Ergodic trajectory generation does not explicitly quantify the uncertainty in the surrounding environment but rather plans a trajectory assuming an information distribution is given. In this work, we use the uncertainty in the factor graph model to guide our exploration.

\textbf{Factor Graphs} are a common way to represent estimation problems for SLAM and other non-linear estimation problems.
A factor graph is a graphical model involving factor nodes and variable nodes.
The variables, or values, represent the unknown random variables in the estimation problem, such as robot poses. 
The factors are probabilistic information on the variables and represent measurements, such as of the signal strength, or constraints between values. 
Performing inference in a factor graph is done through optimization, making it suitable for large scale inference problems~\cite{dellaert_factor_2012}.
Factor graphs have been used for very large-scale SLAM problems~\cite{Lamp2} and for estimation of gas concentrations~\cite{gmonroy_time-variant_2016}.

\textbf{Factor Graph based Informative Path planning}
Factor graphs have been used as a model of a continuous concentration by modeling the problem as a Markov random field. 
This approach has been used to monitor time varying gas distributions in spaces with obstacles by defining a factor graph over the entire space with an unknown value at each cell~\cite{gmonroy_time-variant_2016}. 
It has also been used to jointly estimate the concentration of gas and the wind direction~\cite{gongora_joint_2020}.
These approaches suffer from a scaling problem due to their ties to the geometry of the environment and do not handle unknown environments.

\section{Formulation}
\newcommand{\expectedimprovement}[0]{S}
\newcommand{\actioncost}[0]{C}
\newcommand{\timehorizon}[0]{T_h}
\newcommand{\robotstate}[0]{q}
\newcommand{\action}[0]{a}

\newcommand{\signalstate}[0]{W_s}
\newcommand{\worldstate}[0]{W_r}

In informative path planning for active source seeking, the robot is tasked to localize the source of a radio signal without any prior map of the environment and within limited time constraints.
We define the state as $s=(q, \signalstate, \worldstate)$ where $q$ is the robot state, $\signalstate$ is the signal state (the signal location) and $\worldstate$ is the world traversal risk state.
The task is to find a policy that maximizes an objective function $S$ while minimizing its action cost.
Typically, the objective $S$ corresponds to the expected reduction in uncertainty in the belief over the signal source location $\signalstate$. %
We define the reward for taking action $a$ at state $s$ as
\begin{equation}\label{eq:reward}
R(s,a) = f(\expectedimprovement{}(q' \mid\signalstate) , \actioncost{}(\robotstate,\action \mid W_r)) \quad \text{s.t. } q' = T(\robotstate,\action \mid \worldstate)
\end{equation}
where $C(\robotstate,\action \mid W_r)$ is the cost of taking action $\action$ from state $s$ and $T$ is the robot state transition function, and $f$ is a weighted combination of the two terms defined in PLGRIM~\cite{kim2021plgrim}.

In order to maximize \expectedimprovement, the robot must be able to infer signal strength at unseen locations, given the history of locations $\sensedlocations_{0:t}$ and measurements $\sensedvalues_{0:t}$, as well as hypothetical future locations and measurements, predicted by the model.
This multiple step look-ahead allows the robot to plan a policy which is non-myopic and looks ahead to the reward for complex, multiple step policies.
We denote the belief at time $t$ as $W_s^t = M(\sensedlocations_{0:t},\sensedvalues_{0:t})$ where $M$ is a model of the belief over the signal strength in the environment.
Under this model, we wish to find an optimal policy
\begin{equation}\label{eq:IPPproblem}
    \pi^* = \argmax_{\pi \in \Pi} \sum_{t'=t}^T \mathbb{E}[ R(b_{t'},\pi(b_{t'}))]
\end{equation}
where $t$ is the current planning time, $b_{t'}$ is the belief over state $s$ at time $t'$, and $T$ is the robot's time budget.
As the robot progresses in the mission and updates its world belief, we re-solve the problem in a receding horizon fashion over the remaining time interval $[t,T]$.

\section{Approach}

In this section, we introduce novel signal belief components $\signalstate^g$ and $\signalstate^l$ and propose a hierarchical planner for solving Eq.~(\ref{eq:IPPproblem}).%
We decompose the signal, traversibility, and planning portions into global and local components to allow our system to scale easily through decomposition.

\subsection{World Traversability Belief}\label{sec:world_traversibilty}

Hierarchical planning frameworks address both space complexity (due to the large and increasing environment size) and model uncertainty challenges by breaking down the environment representation into components of different scales and resolutions.
In PLGRIM \cite{kim2021plgrim}, the world belief representation is composed of a meter-level resolution lattice representation of the robot's local surroundings, termed \emph{Local Information Roadmap} (Local IRM), and a topological graph representing the explored space, called \emph{Global Information Roadmap} (Global IRM).
Together, the Local IRM $W_r^l$ and Global IRM $W_r^g$ compose the world traversability belief $W_r$.
This can be seen in \cref{fig:hero} where the local and global information roadmaps are shown next to a traversability map of the environment.

\subsection{Signal Belief}\label{sec:fgs}
\begin{figure}[t]
    \centering
    \begin{subfigure}{.49\columnwidth}
        \centering
        \includegraphics[width=\textwidth]{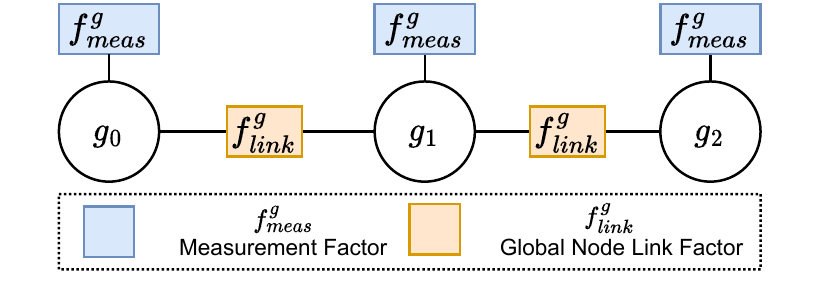}
        \caption{Global factor graph}\label{fig:global_graph}
    \end{subfigure}
    \begin{subfigure}{.49\columnwidth}
        \centering
        \includegraphics[width=\textwidth]{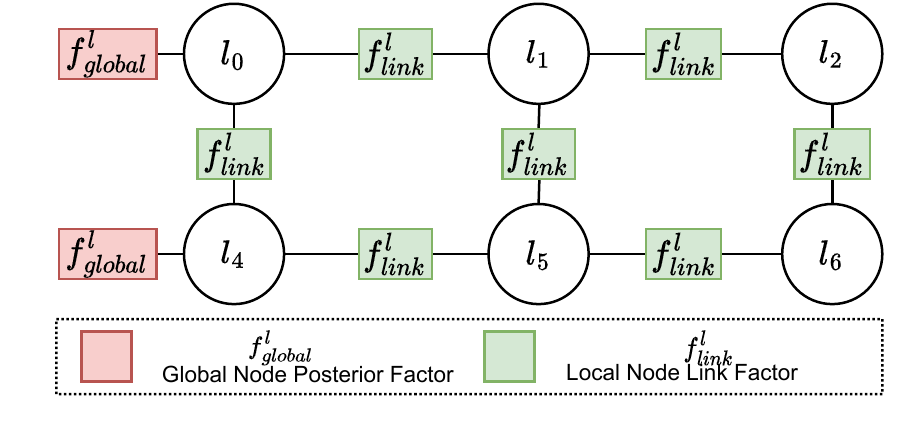} 
        \caption{Local factor graph}\label{fig:local_graph}
    \end{subfigure}
    \caption{{\bf The signal belief is represented by two factor graphs,} described in \cref{sec:fgs}. (a) The global factor graph represents the belief over the signal strength at locations the robot has received measurements, represented as the values $g_1,..,g_n$. (b) The local factor graph is centered at the robot using the values $l_1,...,l_n$ to infer the signal locally around the robot. } 
    \label{fig:my_label}
\end{figure}

\begin{algorithm}[t]
\caption{Add To Global Graph \\
\scriptsize{Add measurement $\sensedvalue_{i}$ at location $\sensedlocation_i$, Given global graph $G^{g}$, Current Estimate $V^{g}$, Distance based signal variance function $\sigma^2_{dist}$, Measurement variance $\sigma^2_{meas}$, Minimum new global node distance $d_{min}$, Global values $g_i, g_{i-1}$ }}\label{alg:add}
\If{$\| \sensedlocation_i - \sensedlocation_{i-1} \| > d_{min}$}{ 
    $f^{g}_{link} \gets  \mathcal{N}(g_i - g_{i-1} \mid 0, \sigma^2_{dist}(\sensedlocation_i, \sensedlocation_{i-1})))$ \\
    
    $f^{g}_{meas} \gets \mathcal{N}(g_i \mid \sensedvalue_i, \sigma^2_{meas}))$ \\
    $G^{g} \gets G^{g} \cup \{f^{g}_{link}, f^{g}_{meas}\} $ \\
}
$V^{g} \gets SOLVE(G^{g}, V^{g})$ \\ \end{algorithm}

In order to facilitate large scale signal inference the model is split into global and local representations of the signal over the environment. 
The local representation in the planner and model represent local beliefs and plans about the area immediately surrounding the robot.
The global representation describes the belief about the signal strength at locations the robot has received measurements.
The posterior distributions over the global signal representation are used as priors in the local signal representation.

\textbf{Global Factor Graph} $(G^g,V^g)$ is seen in \cref{fig:global_graph} where each value $g_n \in V^g$ is connected to at least one measurement factor, $f_{meas} \in G^g$. 
These measurement factors are unary factors which represent the real world measurements taken by the robot at the location $\sensedlocation_n$.
Each global value is connected to the previous and next global value through a link factor, $f^{g}_{link} \in G^g$. 
The link factor connects the previous and current global values, based on distance traveled, while the measurement factor connects the global graph to real world signal measurements.
The global graph does not consider obstacles as the robot only moves a small distance between global node generation.
This link factor, used in both the global and local factor graphs to connect values which have a spatial relationship, serves a similar purpose to the kernel distance in Gaussian processes.
This link factor has an expected mean of 0, with an uncertainty that grows in proportion to the distance the location corresponding to the values are apart $\sigma^2_{dist}(\sensedlocation_i, \sensedlocation_{j}) = \epsilon + \alpha_{dist} \| \sensedlocation_i - \sensedlocation_j \|$ where $\epsilon$ is a small positive number and $\alpha_{dist}$ is an experimentally determined constant. 
In order to add new measurements to the global factor graph, \cref{alg:add} is used, in which a new global value is created only when the robot has moved sufficiently from the current location.
If the robot has not moved far enough from the current location, the another measurement factor is added to the previous value.

\begin{algorithm}[t]
\caption{Create Local Graph \\
\scriptsize{Produces a new local graph $G^{l}$ and local values $V^{l}$, given new local information roadmap $IRM^l$, global graph $G^{g}$, global values $V^{l}$, global k-nearest neighbors function $KNN^g$, number of global nodes to link to $k_{g}$, and robot location $\sensedlocation_t$} }\label{alg:create_local}
$V^{l}, G^{l} \gets \emptyset, \emptyset$ \\
\For{$\sensedlocation_i \in nodes(IRM^l)$}{
    $V^{l} \gets V^{l} \cup \{v_{i}\}$ \\
    \For{$\sensedlocation_j \in neighbor(\sensedlocation_i,IRM^l)$}{
        $\sigma^2_{occ} \gets \sigma^2_{occ} (\sensedlocation_i) + \sigma^2_{occ} (\sensedlocation_j)$ \\
        $f^{l}_{link} \gets \mathcal{N}(l_{i} - l_{j} \mid 0, \sigma^2_{dist}(\sensedlocation_i,\sensedlocation_j) + \sigma^2_{occ})$ \\ 
        $G^{l} \gets G^{l} \cup \{f_{link} \}$ \\
    }
}
\For{$\sensedlocation^{g} \in KNN^{g}(\sensedlocation_t, k_{g})$}{
    $l \gets CLOSEST(x^{g}, V^{l})$ \\
    $\mu(\sensedlocation^g),\sigma^2(\sensedlocation^g) \gets V^{g} (\sensedlocation^g)$ \\
    $f^{l}_{global} \gets \mathcal{N}(l \mid \mu(\sensedlocation^g), \sigma^2(\sensedlocation^g) + \sigma^2_{dist}(l,\sensedlocation^g)$ \\
    $G^{l} \gets G^{l} \cup \{ f^{l}_{global} \}$ \\
}

\end{algorithm}

\textbf{Local Factor Graph} $(G^l, V^l)$ is built from the local information roadmap over the area centered at the robot.
The local information roadmap is an $n \times m$ grid which describes the area the robot can traverse in the local policy, described in \cref{sec:world_traversibilty}.
The local factor graph (\cref{fig:local_graph}) models the local IRM topology by adding link factors between values $l \in V_l$ in the local information roadmap which are connected.
The local link factors, $f^{l}_{link} \in G^l$, are similar to the global link factors as uncertainty increases with the distance between the spatial locations, but also has the addition of occupancy information.
We assume that if a location corresponding to a local value is not traversable for the robot, the model estimate of the signal will have higher uncertainty in that area due to potential attenuation or reflection in the case of solid obstacles.
To model this we increase the uncertainty proportional to the occupancy probability of that location, according to $\sigma^2_{occ}(i) = \alpha_{occ} p_{occ}(i)$, where $\alpha_{occ}$ is an experimentally chosen constant, set to $0.5$ in this work.
In order to incorporate the real world measurements into the local factor graph, we find the closest $k_{g}$ locations corresponding to values in the global factor graph to the robot location. 
For each location that is close to the robot location, we add a factor, $f^{l}_{global} \in G^l$, to the closest local value which uses the posterior estimate from the global value as well as the distance between the local and global locations. We assume the distance is small and do not consider obstacles between the global and local nodes.
The algorithm for this can be seen in \cref{alg:create_local}, which describes how both the local graph and local values are constructed. 

\begin{algorithm}[t]
\caption{Inference at locations \\
\scriptsize{Produces a belief over signal at location $\sensedlocation_q$ given hypothetical future locations $\sensedlocations_h$ and measurements $\sensedvalues_h$, initially empty local value cache $C$, local graph $G^{l}$ and local values $V^{l}$. }}\label{alg:inference}
\uIf{$\sensedlocations_{h} \in C$}{
    \Return $C(\sensedlocations_{h}, \sensedlocation_q) $\\
}
$G_h^{l} = G^{l}$ \\
\For{$(\sensedlocation, \sensedvalue) \in (\sensedlocations_{h}, \sensedvalues_{h})$}{
    $l \gets CLOSEST(\sensedlocation, V^{l})$ \\
    $\mu(l), \sigma^2(l) \gets V^{l}(l)$ \\
    $G_h^{l} \gets G_h^{l} \cup \{ \mathcal{N} (l \mid \sensedvalue,\sigma^2_i) \} $\\
}
$V_h^{l} \gets SOLVE(G_{h}^{l}, V^{l})$ \\
$C(\sensedlocations_{h}) = V_h^{l}$ \\
\Return $V_h^{l}(\sensedlocation_q)$
\end{algorithm}
\textbf{Inference} In order for the planner to do active source seeking, the signal belief must be able to efficiently infer a distribution over the signal at unknown spatial locations. 
The model must also support the ability to condition the model on locations and measurements the robot plans to take, but has not taken yet, termed hypothetical locations $\sensedlocations_h$ and hypothetical values $\sensedvalues_h$.
This allows the planner to make non-myopic plans in which the future values take into account measurements that will be taken in the policy~\cite{marchant_sequential_2014}.
To this end, \cref{alg:inference} describes the algorithm to infer the posterior belief at a spatial location $\sensedlocation_q$. 
Typically, the graph cannot be re-solved with every query so the algorithm determines if it has already computed the local values given the current hypothetical locations $\sensedlocations_h$ by checking if it is in the cache $C$. 
The cache $C$ allows the re-use of previously solved values as the posterior distribution over $\sensedlocation_q$ has already been computed when computing a different query location.
$C$ is emptied when a new local IRM is received due to the robot moving, or a new measurement is added to the global graph.
If algorithm cannot use the cached values, the algorithm adds unary factors representing the conditioning on the hypothetical measurements, using the previous prior uncertainty for the unary factor.
The algorithm then re-solves the factor graph for the entire local information roadmap and adds the newly solved factors to the cache.
This cache is emptied each time a new local factor graph is constructed.

To produce a belief over locations that are not in the local information roadmap, query locations are added to the global roadmap with distance based link factors and the resulting graph is solved. 

\subsection{Hierarchical Solver}

Our approach is to compute a local planning policy that solves \cref{eq:IPPproblem} on $W_r^l$ and $W_s^l$ over a receding horizon $t_l$, which we refer to as \emph{Local Planning},
while simultaneously solving \cref{eq:IPPproblem} over the full approximate world representation $W_r^g$ and $W_s^g$, which we refer to as \emph{Global Planning}.

\textbf{Local Policy} 
In Receding Horizon Planning (RHP), the objective function in Eq.~(\ref{eq:IPPproblem}) is modified:
\begin{align} %
  \pi_{t:t+T}^*(b) &= \argmax_{\pi} \, \mathbb{E} \left[ \sum_{t'=t}^{t+t_l} \gamma^{t'-t} R(b_{t'}, \pi_{t'}(b_{t'})) \right]
  \label{eq:receding_objective_function}
\end{align} where $t_l$ is a finite planning horizon for a planning episode at time $t$.
The expected value is taken over future measurements and the reward is discounted by $\gamma$ \cite{bouman2022adaptive}.

For the local policy, the reward is based on the current signal belief and uses  the upper confidence bound (UCB) acquisition function, which is commonly used for informative path planning for Bayesian optimization~\cite{marchant_sequential_2014}.
The UCB equation is  
    $S^{l}(\sensedlocation_i) = \mu(\sensedlocation_i) + \beta \sigma^2(\sensedlocation_i)$
where $\mu(\sensedlocation)$ and $\sigma^2(\sensedlocation)$ are the current mean and variance of $W_s$ at location $\sensedlocation_i$ and $\beta$ is an empirically determined weight.
We use this objective function for local exploration as the robot should investigate areas that the model is uncertain about, but should still prefer high concentration areas.
In our experiments we set $\beta=3$ to encourage exploration.

\textbf{Global Policy}
Boundaries in between explored and unexplored space, termed \emph{frontiers}, are encoded in the topological graph of the environment (Global IRM) and represent goal waypoints. By visiting frontiers, the robot takes new signal measurements and uncovers swaths of the unexplored environment.
The robot must plan a sequence of frontiers $p$ that it can reach within the time budget that maximizes its expected reward $S^g$. \Cref{eq:IPPproblem} becomes
\begin{equation}
\label{eq:frontloading}
\begin{aligned}
\pi^{\star} = \argmax_{p} \quad \sum_{n \in p} & F \big( 
t_p(n);\worldstate^g) \cdot \expectedimprovement^g \big(n ; \signalstate^g)
\end{aligned}
\end{equation}
where $t_p(n)$ is the time required to reach frontier $n$ through $p$, %
$F$ is the \emph{front-loading} function proposed in \cite{peltzer_fig-op_2022}, and where $p$ should verify the time budget constraint. $F %
\big(t_p(n);\worldstate^g)$ 
is a greedy incentive that encourages gathering reward earlier in time in order to trade off long-term finite-horizon planning with immediate information gain.

We wish to maximize the expected improvement (EI) for signal measurement taken at a frontier $n_i$ (corresponding to location $x_i$), given current signal belief $\signalstate^g$.
EI favors actions that offer the best improvement over the current maximal value, with an added exploration term $\xi$ to encourage diverse exploration.
The EI objective function is defined according to 
$S^{g}(\sensedlocation_i) =  I\Phi(Z) + \sigma(\sensedlocation_i)\phi(Z) $
where $Z =\frac{I}{\sigma^2(\sensedlocation_i)}$,  $I = \mu(\sensedlocation_i) - \max(\mu(\sensedlocations_{0:t})) - \xi$, and $\Phi$ and $\phi$ are the CDF and PDF of the normal distribution, respectively~\cite{jones_efficient_1998, qin_improving_2017}.
Expected improvement is chosen as the global objective function because the robot should prefer frontiers which have a gain in the expected signal strength, while during local exploration the robot should seek out information rich locations rather than only seeking the maxima.

\textbf{Policy Selection} 
The local planning policy $\pi^l$ provides immediate signal reward, and is computed over the local space immediately surrounding the robot, while the global policy $\pi^g$ is computed over a sparse representation of the entire environment the robot has explored so far. Therefore, the global policy has the potential to bring the robot to valuable locations it has not yet explored, but may also require more travel time. %
As in \cite{ott2022riskaware}, we select the planning policy $\pi$ according to its utility $U(b_t; \pi)$ computed from \cref{eq:IPPproblem} and the probability $\hat{P}(\pi)$ that the plan will be successfully executed:
\begin{equation}\label{eq:policyselection}
    \pi_t^\star = \argmax_{\pi \in \lbrace \pi_{t:t+T}^g, \, \pi_{t:t+t_l}^\ell \rbrace}  \hat{P}(\pi) U(b_t; \pi).
\end{equation}

\section{Experiments}
\begin{figure}
    \centering
    \includegraphics[width=\columnwidth]{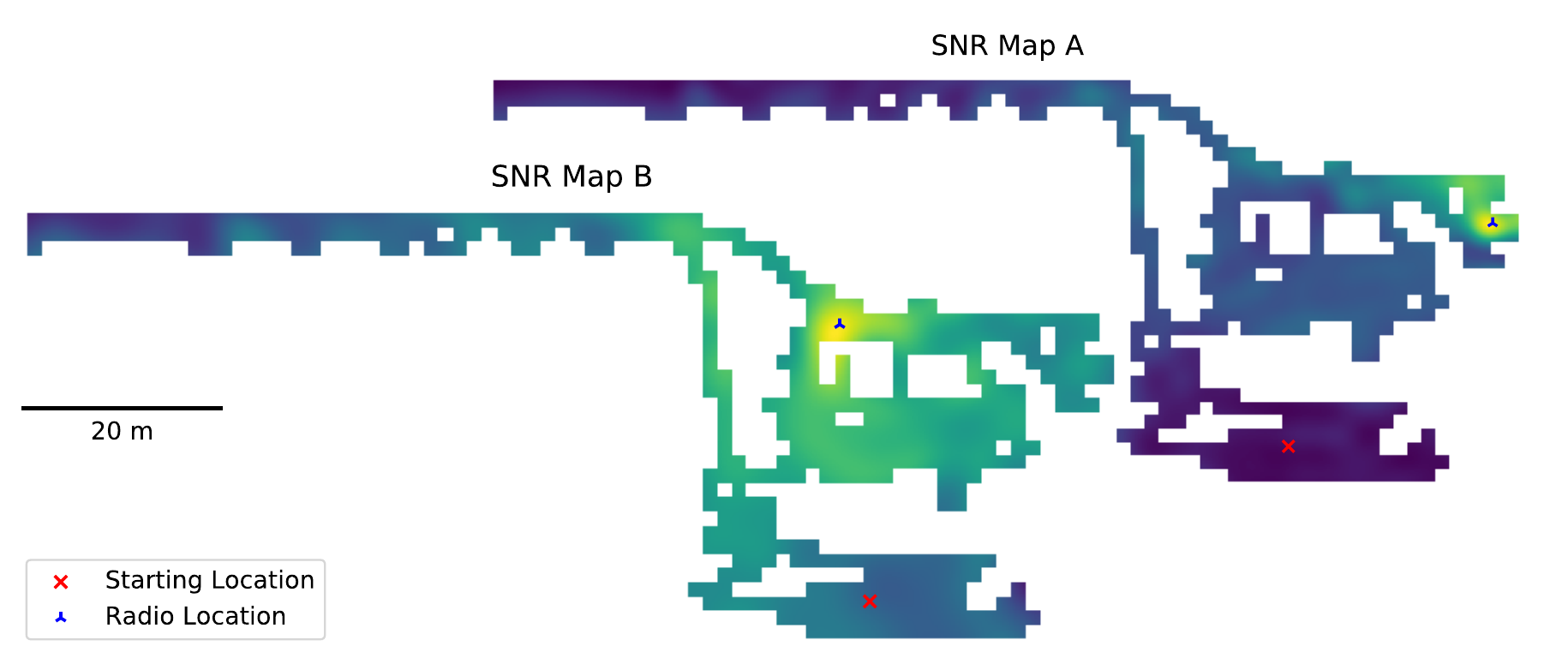}
    \caption{{\bf Maps of the Signal to Noise Ratio collected in a real parking garage.} SNR is colored from blue (low) to yellow (high). Areas in white are not traversable by the robot. 
    }\label{fig:snr_maps}
    \vspace{-1.0em}
\end{figure}
We demonstrate our system on two simulation environments as well as a hardware trial.
The simulated environments were collected in a parking garage with a robot with an attached radio and a radio placed in the environment.
In the simulation environments, we first run the robot in a lawnmower path to collect signal strength data and traversibility data in the environment.
We position the radio in two different locations (\cref{fig:snr_maps}).
Location A is less accessible, so that stronger readings are only seen in direct line of sight, in location B the radio is placed so that it is within line of sight of more areas of the environment.
The robot starts in the same location in both environments, and the traversability is the same in both environment.

In the experiments we compare our approach (labeled Factor graph) with a standard Gaussian process using a squared exponential kernel. The kernel hyper-parameters are set to a length scale of $3$ and an observation noise of $0.01$.
This Gaussian process also only adds new measurements when the robot has moved a minimum distance ($d_{min}$), which was set to $0.25m$ in our experiments.
We also compare against a Gaussian process (GP) which is limited during inference to the same number of measurements ($k_{g}$) during inference as the Factor graph, which we term Gaussian process (limited), (GP (limited)). 
This limited Gaussian process is added to show that the performance of our system is not merely due to limiting the number of measurements included in the model.
For the factor graph, we set the measurement noise to $0.01$,  $\alpha_{occ}$ to $0.5$, and $\alpha_{dist}$ to $0.1$.

\subsection{Simulation Experiments}
\begin{figure}
    \centering
    \begin{subfigure}{0.49\columnwidth}
        \centering
        \includegraphics[width=\textwidth]{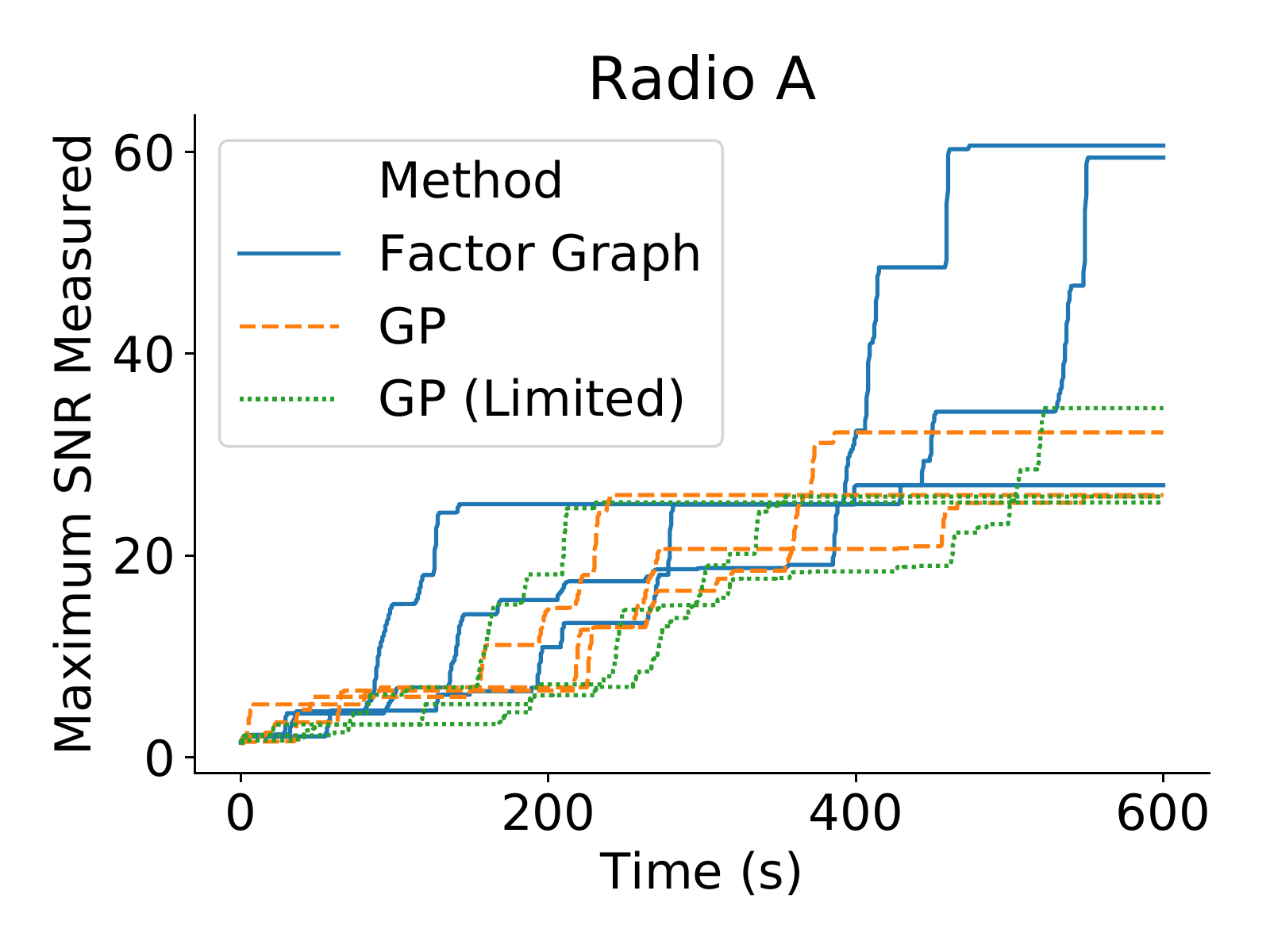}
    \end{subfigure}
    \begin{subfigure}{0.49\columnwidth}
        \centering
        \includegraphics[width=\textwidth]{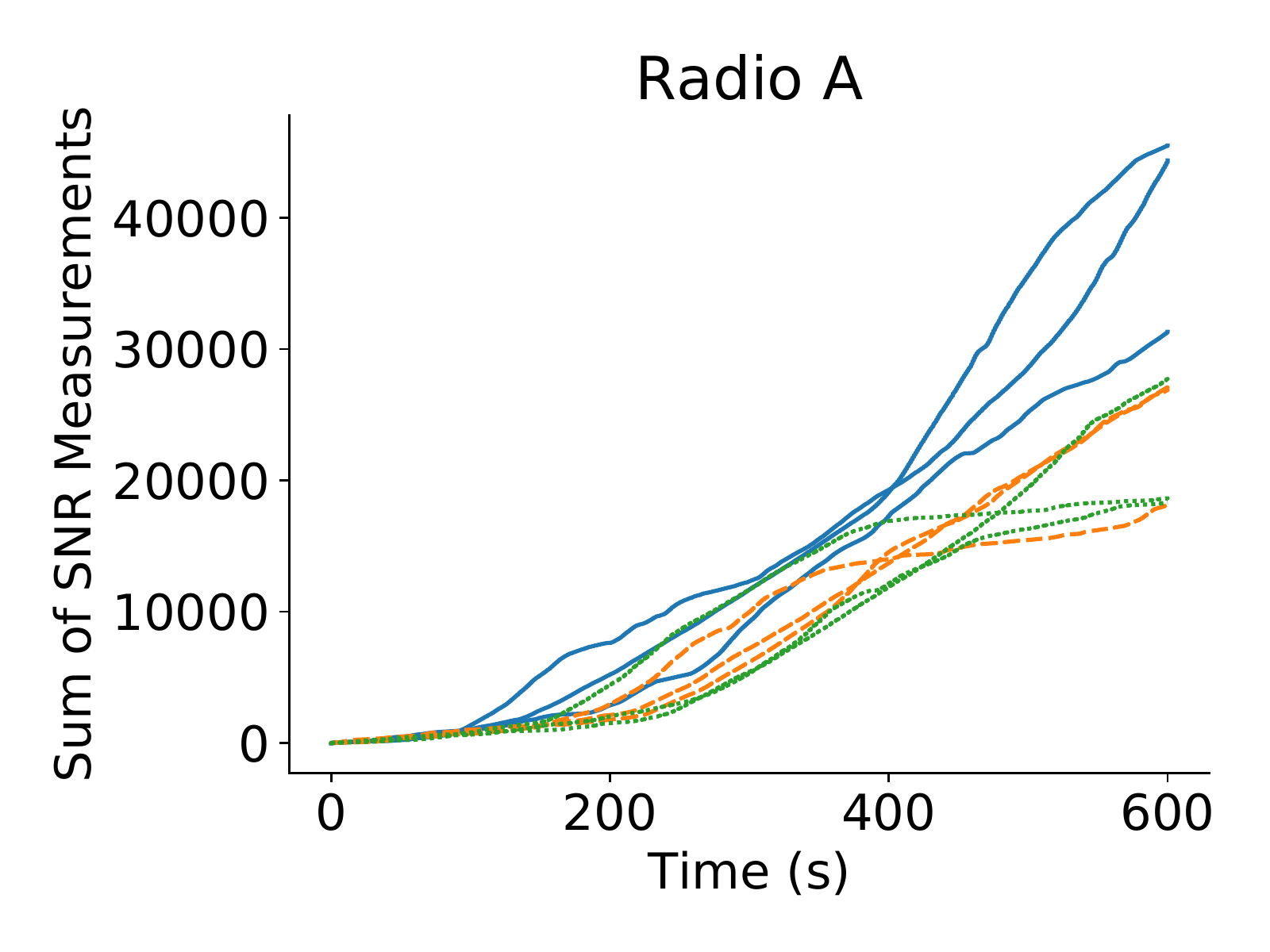}
    \end{subfigure}
    
    \begin{subfigure}{0.49\columnwidth}
        \centering
        \includegraphics[width=\textwidth]{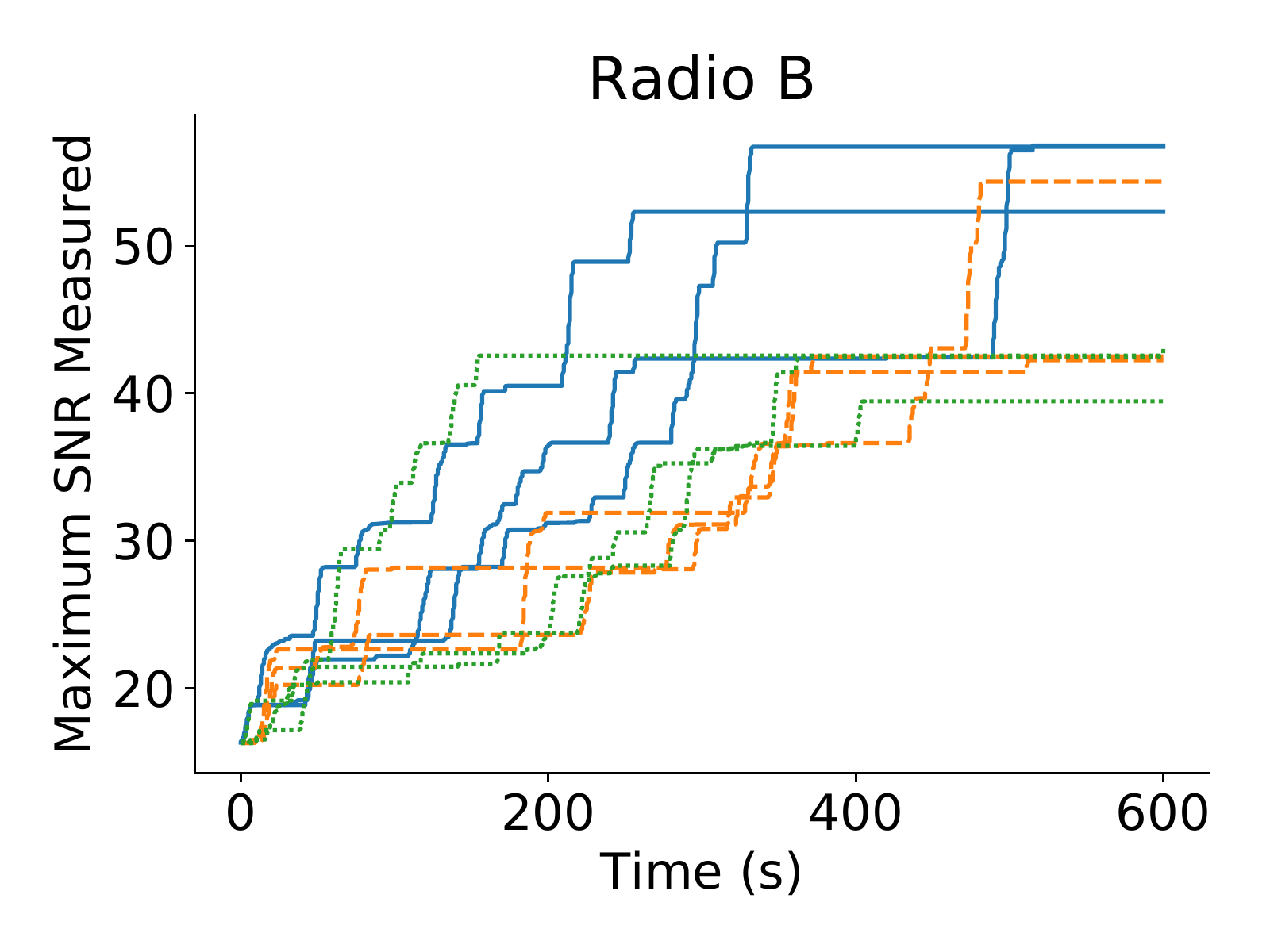}
    \end{subfigure}
    \begin{subfigure}{0.49\columnwidth}
        \centering
        \includegraphics[width=\textwidth]{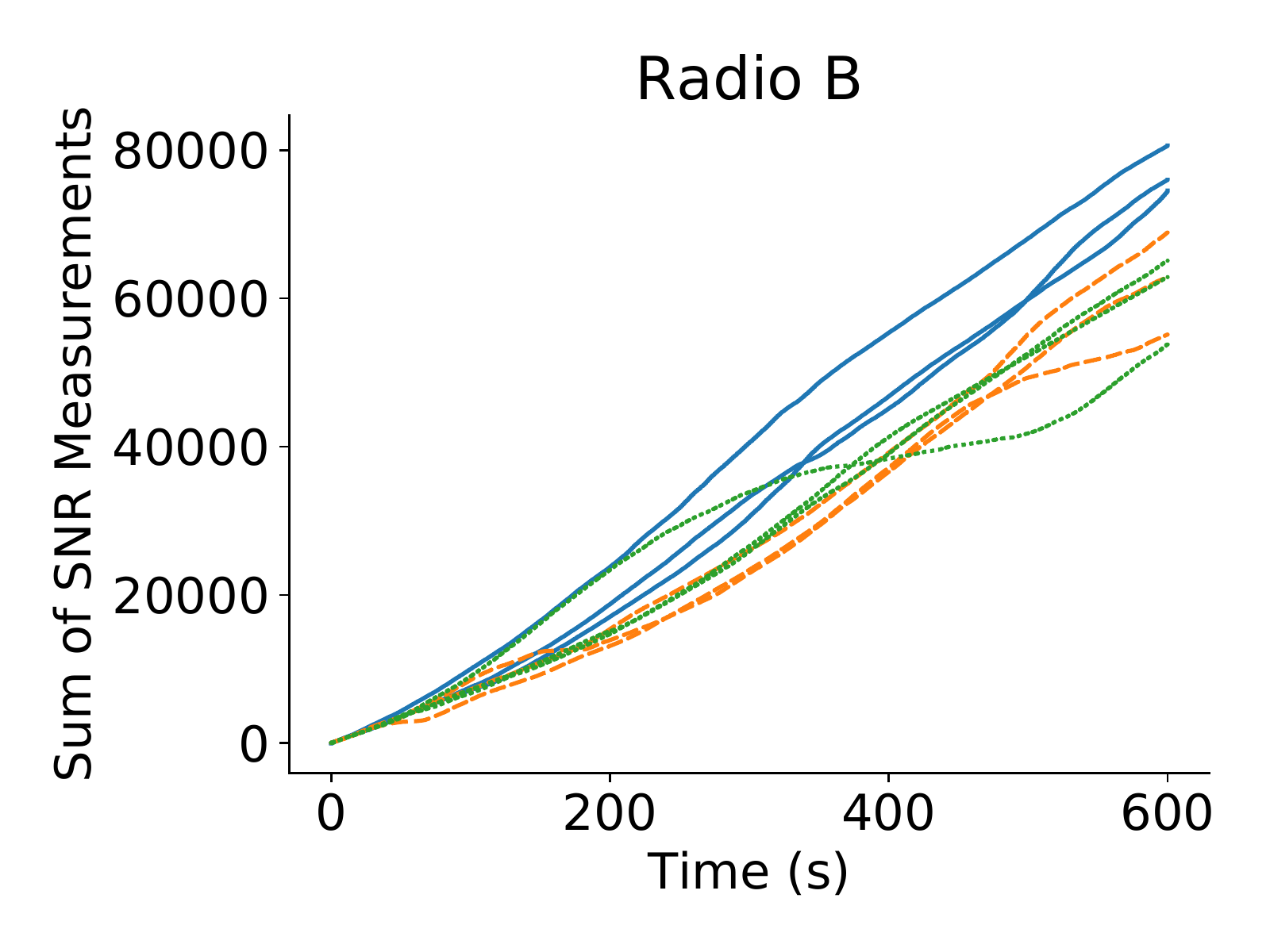}
    \end{subfigure}
    \caption{{\bf Simulation experiment results over two real-world environments.} Left: the maximal signal reading up until time $t$. Right: the sum of the measurements taken until time $t$. }
    \label{fig:sim_results}
    \vspace{-1.5em}
\end{figure}
We compare our novel Factor graph approach with the two baseline approaches (Gaussian Process and Gaussian process (limited)) in both environments (location A and location B).
To compare against the baselines, we show two metrics for each environment (\cref{fig:sim_results}).
The first  shows the maximal SNR measurement recorded up until time $t$. 
It compares how closely the robot approaches the signal source, and how quickly it approaches the signal source.
The second metric is the cumulative sum of the SNR measurements the robot has taken until time $t$. 
This metric, inspired by the cumulative reward metric~\cite{marchant_sequential_2014}, shows how the robot makes decisions to find maximal areas over time.
For each environment, we run the robot three times and report each trial.

As can be seen in \cref{fig:sim_results}, the factor graph model allows the robot to find a much higher signal strength in 2 out of 3 trials, and the performance is more consistent. 
Limiting the Gaussian process to only 200 measurements does not have a big impact on the performance compared to the regular Gaussian process, but it still does not perform as well as the factor graph model.
In the comparison of the cumulative signal measurement the factor graph outperforms the other methods in all 3 trials, with the other two methods performing about as well.
This implies the factor graph consistently allows the planner to make better decisions through the entire experiment.

\begin{figure}[t]
    \centering
    \begin{subfigure}{0.43\columnwidth}
        \centering 
        \includegraphics[width=\textwidth]{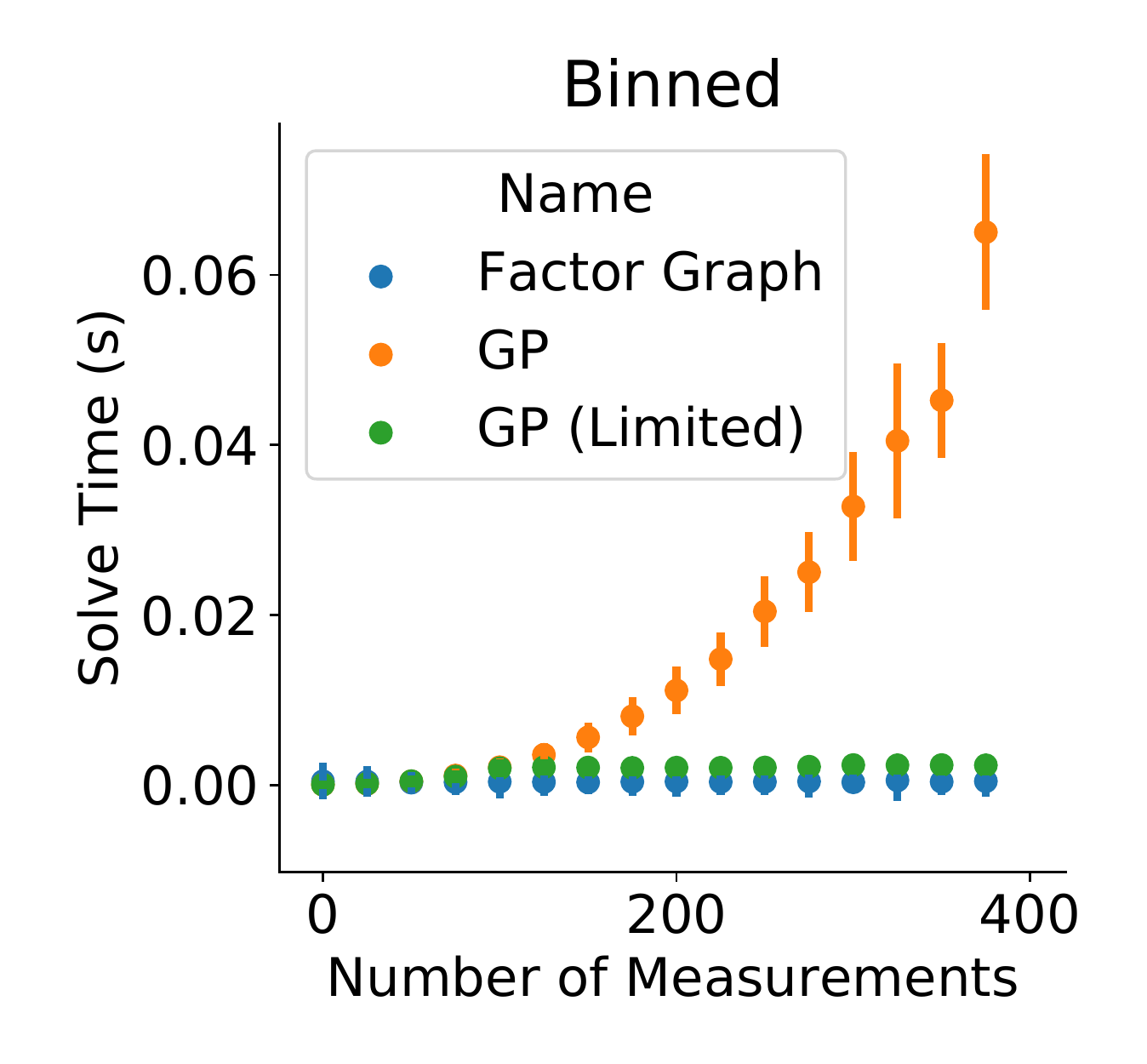}
    \end{subfigure}
    \begin{subfigure}{0.555\columnwidth}
        \centering 
        \includegraphics[width=\textwidth]{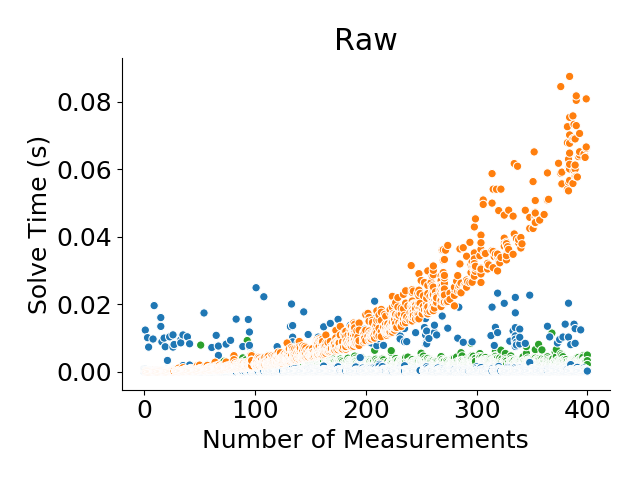}
    \end{subfigure}
    \caption{{\bf Comparison of time required to perform inference for the three models over the number of measurements in the model.} Left: the data binned data with the standard deviation, Right: the raw data. 
    }\label{fig:scaling}
    \vspace{-1.45em}
\end{figure}

We also present a study of the scalabilty of the approaches, shown in \cref{fig:scaling}.
We find that the Gaussian process has the typical exponential scaling with in the time required for inference as the number of measurements in the model grows.
We also find that even limiting the Gaussian process to a fixed number of measurements per inference is still is more expensive than the factor graph on average.
The factor graph has a bimodal distribution in its performance, when the graph has to be re-solved it is slower than the Gaussian process methods, but due to its ability to cache the model posterior, most of the inference is a hash table lookup.

We believe the consistent performance of the factor graph is due to to two reasons. 
The first is that the factor graph is generally faster for inference than Gaussian process methods, allowing the planner to plan at a higher frequency.
The second reason is the ability of the factor graph to represent uncertainty in the signal due to obstacles. 
This allows the factor graph model to infer that signal strengths on the opposite side of obstacles and around corners are more interesting as the signal uncertainty increases through the obstacle.

\subsection{Hardware Experiments}\label{sec:hardware}
\begin{figure}[t]
    \centering
    \includegraphics[trim={1.0cm 0.8cm 0.1cm 0.1cm},clip,width=\columnwidth]{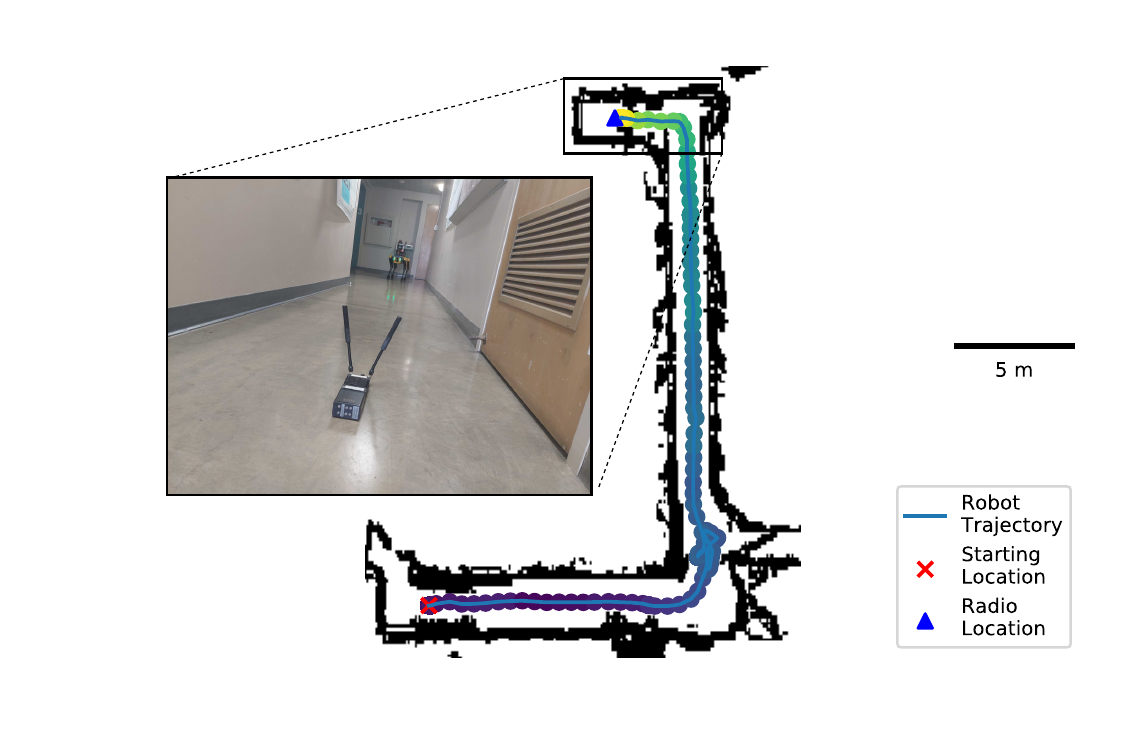}
    \caption{{\bf Experiment of our system on a real Spot robot.} The robot explores a man made structure to find the radio. The collected measurements, colored from blue (low) to yellow (high), as well as the robot's trajectory are shown. Black dots show the non-traversable areas such as walls.}
    \label{fig:demo_expr}
    \vspace{-1.0em}
\end{figure}

We demonstrate our system on a real robot locating a radio in an unknown environment.
The robot, a Boston Dynamics Spot~\cite{spot} equipped with custom sensing and computing systems~\cite{agha2021nebula,Otsu2020,bouman2020autonomous}, is initially started at one end of a parking garage.
The radio is deployed at another end of the hallway after two lefthand turns.
We use the factor graph model as with $k_{g}$ equal to 100 due to the fact that the computer payload on the robot is limited.
As can be seen in \cref{fig:demo_expr}, the robot explores an open passage before going directly to the source of the radio.
We find that our system is able to be easily tuned to the computational limits of the robot and efficiently finds the signal source.
As with \cref{fig:scaling} we found that the robot does not increase its planning time drastically when more measurements are added to the model.
\iros{
An extended hardware experiment can be seen in~\cite{arxiv_version}.
}

\section{Conclusion}
Large scale active source seeking using a robot is a complex and challenging task which requires the ability to scale the planner and model.
In this work we have demonstrated the ability of our factor graph based model to decompose the problem into local and global tasks which allows the model to scale easily with a large number of measurements. 
We have also demonstrated the model's ability to handle the propagation of signals through unknown obstacles by modeling the explicit uncertainty due to these obstacles.
Through simulation experiments we have shown that the ability to plan rapidly due to the inference speed of the model as well as the ability to represent signal uncertainty due to obstacles allow our overall system to perform active source seeking better than the baseline methods.
We have also shown the ability of our model to be scaled to the computational needs of a real life robot and demonstrated our system in a real life unknown environment.

\arxiv{
\section{Appendix}
\subsection{Extended Hardware Experiment}

\begin{figure}[t]
    \centering
    \begin{subfigure}{\columnwidth}
    \includegraphics[trim={0.1cm 0.8cm 0.1cm 0.1cm},clip,width=\columnwidth]{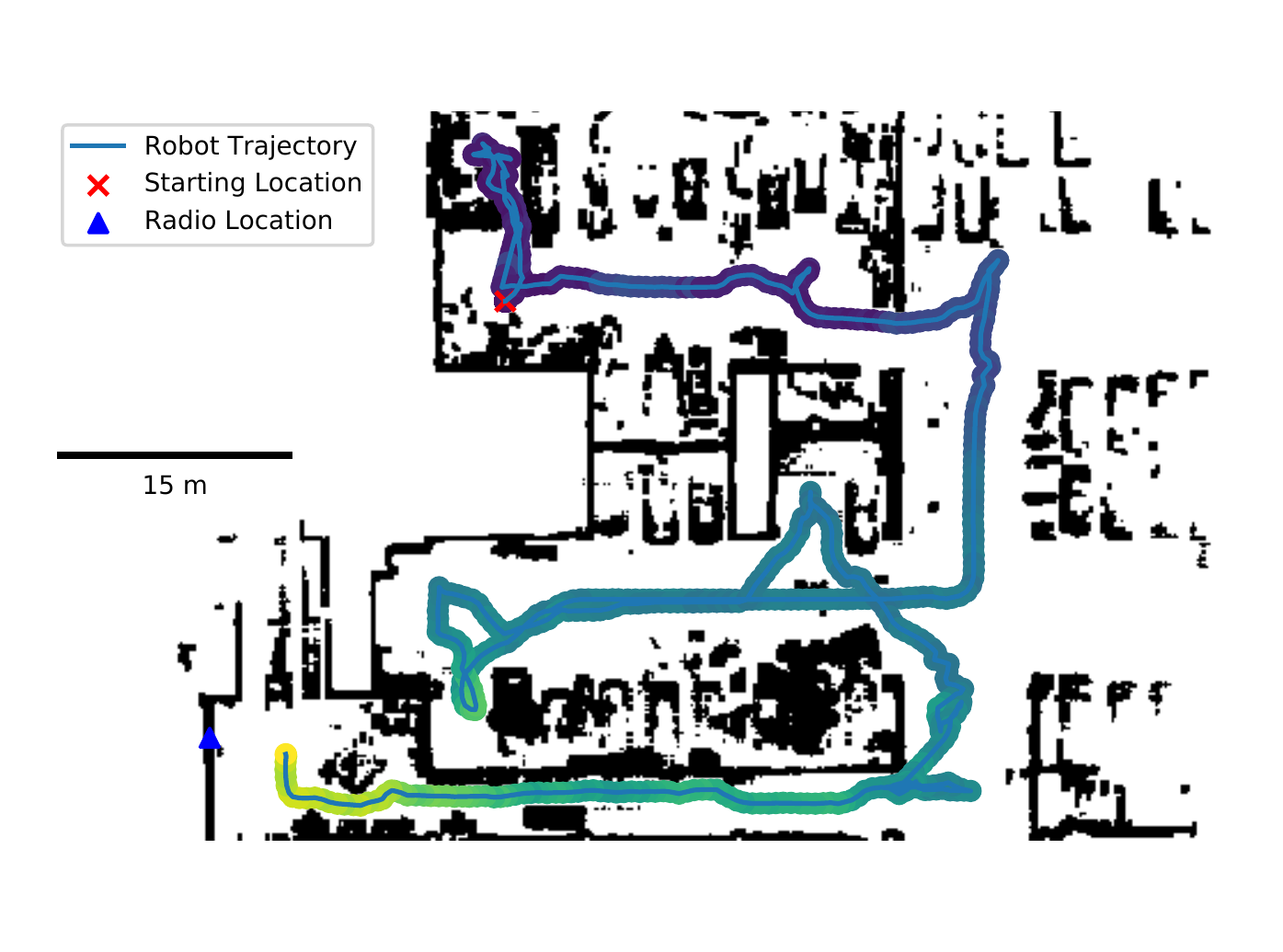}
    \end{subfigure}
    \begin{subfigure}{\columnwidth}
    \includegraphics[width=\textwidth]{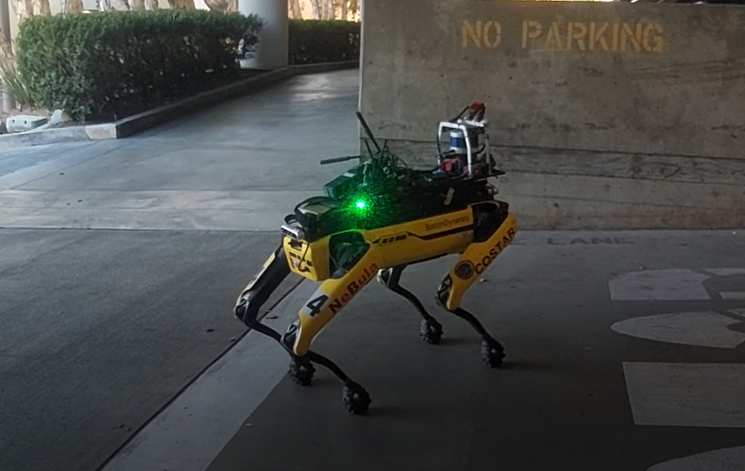}
    \end{subfigure}
    \caption{{\bf Experiment of our system on a real Spot robot in extended field conditions.} Top: The robot explores a man made structure to find the radio. The collected measurements, colored from blue (low) to yellow (high), as well as the robot's trajectory are shown. Black dots show the non-traversable areas such as walls and cars.
    Bottom: Spot robot platform used in this experiment and the experiment in \cref{sec:hardware}. }
    \label{fig:ext_demo_expr}
    \vspace{-1.0em}
\end{figure}

We present an extended study of the performance of our system in the field. 
The setup is similar to \cref{sec:hardware}, including robot platform.
The robotic platform is limited not only in computational power, but also further limited due to the multiple other systems which must run in parallel on the robot to allow for localization, control, and high level mission objectives.
To allow for rapid re-planning, we sub-sample the local factor graph, discarding every other node.
The environment is a parking garage with the target source placed outside of the garage.

As can be seen in \cref{fig:ext_demo_expr}, the robot exits the first corridor, attempts to go down the second corridor but discovers that it cannot directly go to the source in that corridor.
It then exits the second corridor, attempting to discover an unseen passage, then goes to the third corridor to go directly to the source. 

This demonstrate the ability of our system to guide the robot to the signal source in the presence of limited computation and difficult real-world environments. 

}
\printbibliography
\end{document}